\documentclass{article}

\PassOptionsToPackage{numbers}{natbib}

\usepackage[preprint]{neurips_2023}

\usepackage[utf8]{inputenc} 
\usepackage[T1]{fontenc}    
\usepackage{hyperref}       
\usepackage{url}            
\usepackage{booktabs}       
\usepackage{amsfonts}       
\usepackage{nicefrac}       
\usepackage{microtype}      
\usepackage{xcolor}         
\usepackage{amsmath,amssymb,amsfonts,bm}
\usepackage{makecell}
\usepackage{graphicx}
\usepackage{multirow}
\usepackage{chngcntr}
\usepackage{subcaption}

\title{Inverse Probability of Treatment Weighting with Deep Sequence Models Enables Accurate treatment effect Estimation from Electronic Health Records}

\author{%
  Junghwan Lee$^1$ \quad Simin Ma$^1$ \quad Nicoleta Serban$^1$ \quad  Shihao Yang$^{1,*}$\\
  $^1$Georgia Institute of Technology\\
  $^*$Corresponding Author \\
  \texttt{\{jlee3541,sma318\}@gatech.edu} \\
  \texttt{\{nicoleta.serban,shihao.yang\}@isye.gatech.edu}
}

\begin{document}

\maketitle

\begin{abstract}

Observational data have been actively used to estimate treatment effect, driven by the growing availability of electronic health records (EHRs). However, EHRs typically consist of longitudinal records, often introducing time-dependent confoundings that hinder the unbiased estimation of treatment effect. Inverse probability of treatment weighting (IPTW) is a widely used propensity score method since it provides unbiased treatment effect estimation and its derivation is straightforward. In this study, we aim to utilize IPTW to estimate treatment effect in the presence of time-dependent confounding using claims records. Previous studies have utilized propensity score methods with features derived from claims records through feature processing, which generally requires domain knowledge and additional resources to extract information to accurately estimate propensity scores. Deep sequence models, particularly recurrent neural networks and self-attention-based architectures, have demonstrated good performance in modeling EHRs for various downstream tasks. We propose that these deep sequence models can provide accurate IPTW estimation of treatment effect by directly estimating the propensity scores from claims records without the need for feature processing. We empirically demonstrate this by conducting comprehensive evaluations using synthetic and semi-synthetic datasets. 

\end{abstract}

\section{Introduction}

Randomized controlled trials (RCTs) are considered the gold standard to estimate treatment effect~\cite{hariton2018randomised, sibbald1998understanding}. However, RCTs are not always feasible due to high experimental cost, time constraints, and ethical considerations~\cite{soni2019comparison,zeng2022uncovering}. Because observational data are less constrained by the limitations inherent in RCTs, they have become more commonly used in studies aiming to estimate treatment effect~\cite{olier2023causal}. Electronic health records (EHRs) are an important type of observational data that contain rich information about the medical history of an individual~\cite{adler2017electronic}. With the increasing availability of EHRs in the healthcare domain~\cite{evans2016electronic}, EHRs with different types of data modalities, such as time series physiological measurements, claims records, and clinical notes, have also been used for treatment effect estimation~\cite{shi2022learning, zeng2022uncovering}. While utilizing EHRs can mitigate the limitations of RCTs, they often contain confoundings that hinder the unbiased estimation of treatment effect.

\textit{Confounding variables}, also known as \textit{confounders}, are variables that influence both treatment assignment and outcome~\cite{jager2008confounding}. Confounding is present in a study when it includes such variables. EHRs usually contain medical records arranged in chronological order, leading to having \textit{time-dependent confounding}. Adjusting for confounding is essential to accurately estimate treatment effect, but it can be particularly challenging with time-dependent confounding since time-dependent confounding depends on the temporal pattern and dependencies of the confounding variables that change over time and are affected by past features.

Propensity score methods are commonly used to adjust for confounding in treatment effect estimation. These methods leverage the balancing property of the propensity scores, which ensures that the distribution of observed covariates is identical between treated and untreated groups when conditioned on the propensity score~\cite{rosenbaum1983central}. Inverse probability of treatment weighting (IPTW) is a propensity score method that creates a synthetic sample where treatment assignment is independent of the observed covariates by weighting the sample with the inverse of its propensity score. IPTW is widely used since it provides unbiased treatment effect estimation and its derivation is straightforward~\cite{hirano2003efficient,austin2015moving}.

Claims records are a subset of EHRs, containing longitudinal records of standardized codes for diagnoses, medications, and medical procedures. They are a crucial data source for research in medicine and healthcare domains~\cite{shi2022learning}. While claims records are often used to estimate treatment effect with adjustments for confounding using propensity score methods~\cite{seeger2005application, ross2021veridical}, there is a lack of studies focusing on estimating treatment effect using claims records in the presence of time-dependent confounding.

\begin{figure}[t!]
\centering 
\begin{subfigure}[h]{0.45\linewidth}
\includegraphics[width=\linewidth]{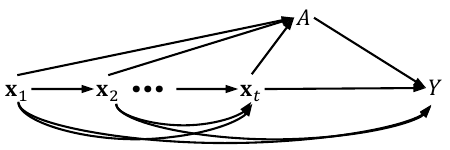}
\caption{}
\label{fig:causal_diagram_1}
\end{subfigure}
\hspace{+30pt}
\begin{subfigure}[h]{0.45\linewidth}
\includegraphics[width=\linewidth]
{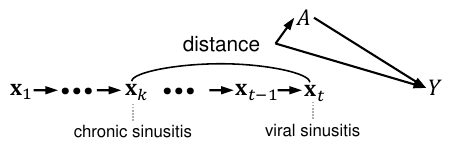}
\caption{}
\label{fig:causal_diagram_2}
\end{subfigure}
\caption{(a) Causal diagram of our problem setup. $A$ denotes binary treatment and $Y$ denotes continuous outcome. A claims record $\mathbf{x}_t$ includes medical codes and also can include a treatment assignment $A$. (b) Causal diagram of a hypothetical confounding scenario in our experiment using a semi-synthetic dataset. The confounding depends on the record-wise distance between \textit{chronic sinusitis} and \textit{viral sinusitis}. Arrows between records are omitted for readability.}
\label{fig:causal_diagram}
\end{figure}

In this study, we aim to utilize IPTW to estimate treatment effect in the presence of time-dependent confounding using claims records. IPTW can provide an unbiased estimation of treatment effect, but it requires an accurate estimation of the propensity scores to ensure this unbiasedness. The challenge lies in accurately estimating the propensity scores in the presence of time-dependent confounding, as it involves capturing the temporal patterns and dependencies of confounding variables across longitudinal records. For example, Figure~\ref{fig:causal_diagram_2} depicts our hypothetical scenario of time-dependent confounding in the experiment using a semi-synthetic dataset, where the confounding depends on the record-wise distance between the code of \textit{chronic sinusitis} and \textit{viral sinusitis}. Previous studies have utilized propensity score methods with features derived from claims records through feature processing. However, such feature processing usually requires domain knowledge and additional resources to extract information to accurately estimate propensity score~\cite{schneeweiss2009high}. 

Deep sequence models, such as recurrent neural networks (RNNs) and self-attention-based architectures, have demonstrated impressive performance in modeling sequential data~\cite{sutskever2014sequence,vaswani2017attention} and have been successfully adopted for modeling EHRs in various tasks~\cite{xiao2018opportunities,mcdermott2021comprehensive,siebra2024transformers}. We propose that deep sequence models based on these architectures can provide accurate IPTW estimation of treatment effect by directly estimating the propensity scores from claims records, even in the presence of time-dependent confounding, without the need for feature derivation from the EHRs. We empirically demonstrate this by conducting comprehensive evaluations, comparing baseline methods that use feature processing with deep sequence models using synthetic and semi-synthetic datasets. For the baseline methods, we selected logistic regression and multi-layer perceptions, the two most widely used methods for propensity score estimation with feature processing. Our results show that IPTW with deep sequence models provides better estimates of treatment effect compared to baseline methods, without requiring additional feature processing.

\section{Methods}

\subsection{Preliminaries}
\label{subsection:prelim}



\paragraph{Potential Outcome framework and average treatment effect} Consider a binary treatment setting with two possible treatments: treated ($A=1$) and control ($A=0$). The potential outcome framework assumes each individual $i$ has a pair of potential outcomes $Y^{(i)}_{A=1}$ and $Y^{(i)}_{A=0}$. For notational simplicity, we denote these potential outcomes as $Y^{(i)}_{1}$ and $Y^{(i)}_{0}$, respectively. The average treatment effect (ATE) is defined as follows:
\begin{equation}
\label{eq:potential_outcome}
    \mathbb{E} \left [ Y_1 - Y_0 \right ] = 
    \mathbb{E} \left [ Y_1 \right ] - \mathbb{E} \left [ Y_0 \right ].
\end{equation}
However, ATE cannot be directly calculated from equation (\ref{eq:potential_outcome}) since only one of the potential outcomes, either $Y_1$ or $Y_0$, can be observed for each individual~\cite{rubin2005causal,rubin1974estimating}. When treatment assignment is random, we know that $\mathbb{E}\left [Y | A = 1  \right ] = \mathbb{E} \left [ Y_1\right ]$ and $\mathbb{E}\left [Y | A = 0  \right ] = \mathbb{E} \left [ Y_0 \right ]$. Therefore, under randomized treatment assignment, ATE can be calculated as:
\begin{equation}
\label{eq:ate_randomization}
    \mathbb{E} \left [ Y_1 - Y_0 \right ] = 
    \mathbb{E} \left [ Y | A = 1 \right ] - \mathbb{E} \left [ Y | A = 0 \right ].
\end{equation}
The ATE calculation does not generally hold in observational studies where randomization of treatment assignment cannot be guaranteed and confounding exists.

\paragraph{Propensity score.} Propensity score, which indicates the probability of an individual being assigned to treatment given the observed covariates of the individual~\cite{rosenbaum1983central}, is defined as follows:
\begin{equation}
e(x) = P(A=1| x ),
\label{eq:propensity_score}
\end{equation}
where $x$ is a covariate vector. In observational studies, the true propensity score is typically unknown and needs to be estimated:
\begin{equation}
    \hat{e} = f_{\theta} (x) ,
    \label{eq:propensity_score_est}
\end{equation}
where $f_{\theta}$ is a function with a parameter set $\theta$ that maps $x$ to the estimated propensity score $\hat{e}$.

\paragraph{Average treatment effect estimation using inverse probability of treatment weighting.} In this study, we use inverse probability of treatment weighting (IPTW) to estimate the ATE. The ATE can be estimated using IPTW as follows:
\begin{equation}
\hat{\Delta}_{\rm{IPTW}} = \frac{1}{N} 
\left ( \sum_{i=1}^{N} \frac{A^{(i)} Y^{(i)}}{\hat{e}^{(i)}} -
\sum_{i=1}^{N} \frac{(1-A^{(i)}) Y^{(i)}}{(1-\hat{e}^{(i)})} \right ),
\label{eq:ate_iptw}
\end{equation}
where $N$ is the total number of samples in the dataset and $\hat{e}^{(i)}$ is the estimated propensity score of individual $i$. Based on the potential outcome framework, $\hat{\Delta}_{\rm{IPTW}}$ in equation (\ref{eq:ate_iptw}) is an unbiased estimation of the ATE~\cite{hirano2003efficient,lunceford2004stratification}. Throughout this paper, we refer to the estimated ATE using IPTW, calculated by the equation (\ref{eq:ate_iptw}), as the estimated ATE.

\subsection{Problem setup}

Figure~\ref{fig:causal_diagram_1} depicts the causal diagram of our problem setup. We consider a claims records dataset $\mathcal{D}= \left \{ \{ \mathbf{x}_t^{(i)} \}_{t=1}^{T^{(i)}}   \cup \{ {A^{(i)}}, Y^{(i)} \} \right \}_{i=1}^{N}$, which consists of records from $N$ independent samples. Each sample $i$ contains records at $T^{(i)}$ discrete time points. We assume that a binary treatment $A \in \{ 0,1 \}$ and a continuous real-valued outcome $Y \in \mathbb{R}$ are observed for each sample $i$ at the observation end time point. We can view each record as a bag of medical codes that can be represented using multi-hot encoding. For instance, a record containing medical codes for \textit{fever} and \textit{acetaminophen} can be expressed as a bag with \textit{fever} and \textit{acetaminophen}. Each record can then be represented as a multi-hot encoded vector, where the elements for these codes are ones, and the rest are zeros. Therefore, for each sample $i$, we observe a record $\mathbf{x}_t^{(i)} \in \{ 0,1 \}^{d_x}$ at time $t$, where $d_x$ denotes the dimensionality of the records (i.e., the number of unique medical codes).

Our objective is to estimate the ATE using IPTW as defined in the equation (\ref{eq:ate_iptw}) on the claims records dataset $\mathcal{D}$ that may contain time-dependent confounding. Specifically, we seek to obtain a model $f_{\theta}$ that maps $X^{(i)}$ to the estimated propensity score $\hat{e}^{(i)}$ as:
\begin{equation}
    \hat{e}^{(i)} = f_{\theta} (X^{(i)}),
    \label{eq:problem_setup}
\end{equation}
where $X^{(i)}$ is the set of records for sample $i$. A desirable model would be able to estimate $\hat{e}^{(i)}$ as closely as possible to the true propensity score $e^{(i)}$, thereby providing an accurate estimation of the ATE using IPTW.

\begin{figure}[t!]
\centering 
\begin{subfigure}[h]{0.6\linewidth}
\includegraphics[width=\linewidth]{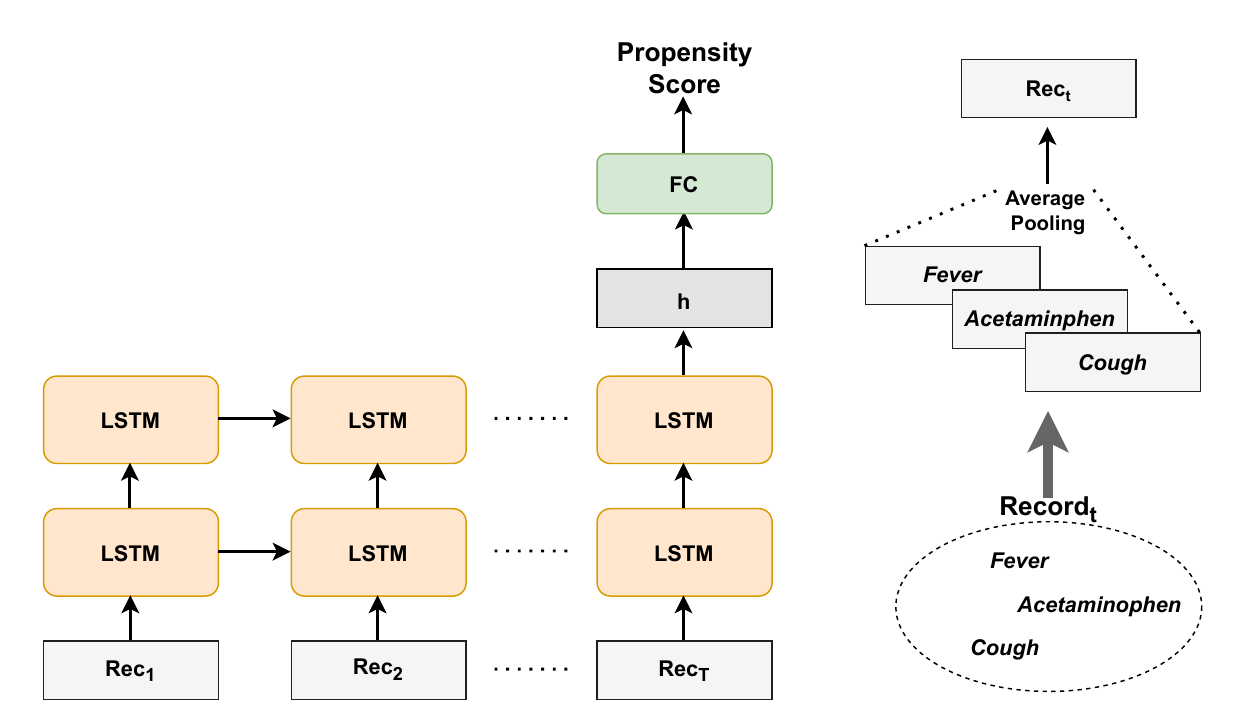}
\caption{LSTM}
\label{fig:LSTM}
\end{subfigure}
\hspace{+30pt}
\begin{subfigure}[h]{0.45\linewidth}
\includegraphics[width=\linewidth]
{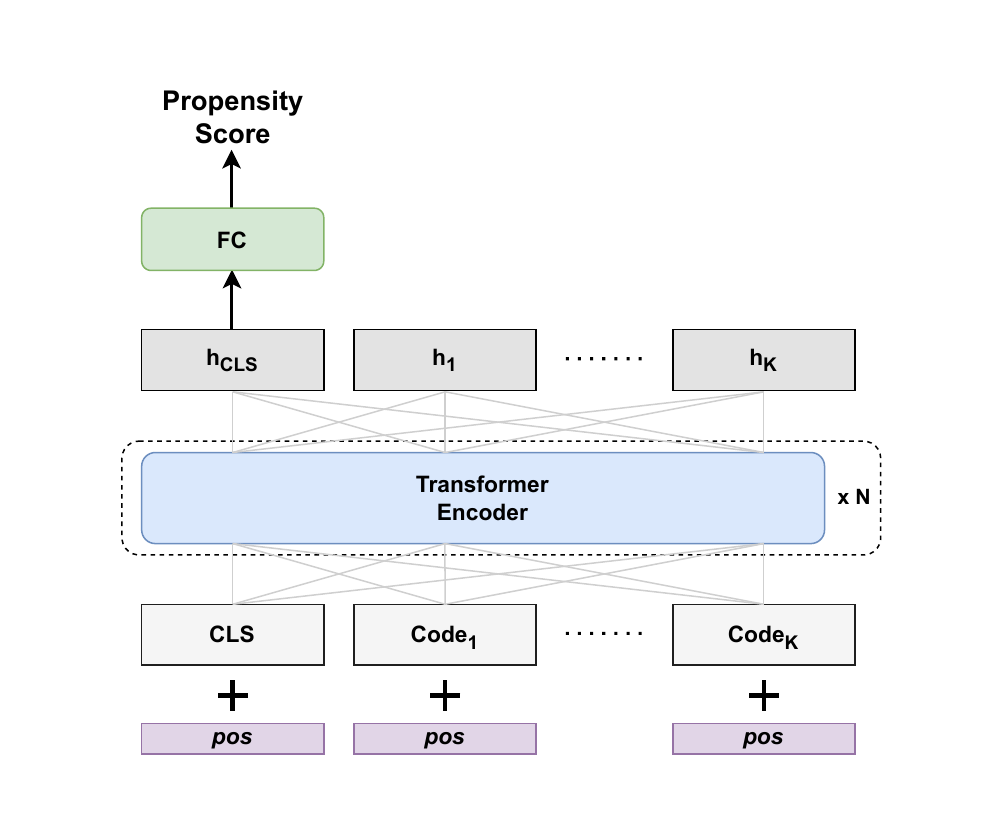}
\caption{$\text{BERT}_{\rm code}$}
\label{fig:BERT_code}
\end{subfigure}
\begin{subfigure}[h]{0.45\linewidth}
\includegraphics[width=\linewidth]
{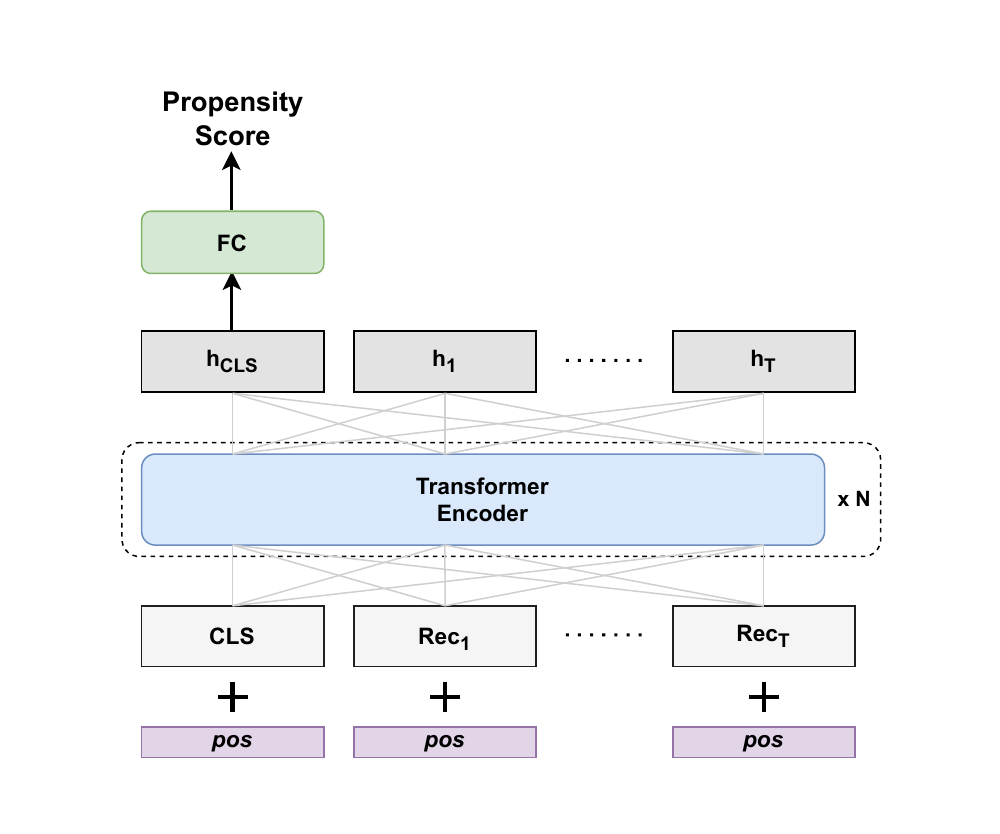}
\caption{$\text{BERT}_{\rm record}$}
\label{fig:BERT_record}
\end{subfigure}
\caption{(a) LSTM to estimate propensity score using claims records. Average pooling is applied to the code representations to generate record representation, aggregating the representations of the codes present in the record. For example, if $\text{Record}_t$ contains \textit{Fever}, \textit{Acetaminophen}, and \textit{Cough} codes, the record representation of $\text{Record}_t$ is generated by averaging the representations of these three codes. (b) $\text{BERT}_{\rm code}$ to estimate propensity score using claims records. The input representations are constructed by arranging code representations in chronological order. (c) $\text{BERT}_{\rm record}$ to estimate propensity score using claims records. The input representations are constructed using record representations, similar to LSTM.}
\label{fig:model_architectures}
\end{figure}

\subsection{Recurrent Neural Networks}

Recurrent neural networks (RNNs) are commonly used deep sequence model architectures with a long history of being used to model sequential data. RNNs have shown significant success in various tasks, such as machine translation~\cite{sutskever2014sequence} and speech recognition~\cite{graves2013speech}. Due to the sequential nature of EHRs, RNNs have also been actively used in modeling EHRs for various medical tasks. For example, Lipton \textit{et al.}  \cite{lipton2015learning} used Long Short-Term Memory Networks (LSTM)~\cite{hochreiter1997long} and Choi \textit{et al.,}~\cite{choi2016doctor} used Gated Recurrent Units (GRU)~\cite{cho2014learning} to predict medical codes of patients in intensive care units, respectively. Additionally, Lee \textit{et al.} \cite{lee2021severity} used GRU to predict the mortality of COVID-19 patients.

We employ LSTM as the model $f_{\theta}$ to estimate the propensity scores in accordance with our problem setup. Figure~\ref{fig:LSTM} visually illustrates how LSTM estimates the propensity scores using claims records. Initially, we apply average pooling to obtain a representation for each record, which aggregates the representations of the codes present in the record (i.e., codes in the bag of records). Code representations are generated using an embedding layer. The record representations are then sequentially fed into the LSTM. A fully-connected neural network layer with sigmoid activation is applied to the final hidden state output by the LSTM to estimate the propensity scores. The model is trained by minimizing the cross-entropy loss between the estimated propensity scores and the binary treatment assignment.

\subsection{Stacked Transformer Encoder Layers}

Transformer~\cite{vaswani2017attention} demonstrated remarkable performance in machine translation and has become the model of choice in various natural language processing (NLP) tasks. The original Transformer architecture is composed of an encoder and a decoder designed for machine translation. The encoder, composed of multiple identical encoder layers, maps an input sentence to a continuous representation. Similarly, the decoder, consisting of multiple identical decoder layers, transforms the input representations into output representations, which are then converted into the desired language. Building upon the Transformer model, encoder-only and decoder-only architectures were developed by stacking multiple Transformer encoder and decoder layers. Notably, BERT (Bidirectional Encoder Representations from Transformers)~\cite{devlin2019bert} and GPT (Generative Pre-Training)~\cite{radford2019language} are pioneering models based on encoder-only and decoder-only architectures, respectively. BERT, in particular, has been widely used in modeling EHRs for various medical downstream tasks, employing a transfer learning scheme similar to that used in the NLP domain, where the model is pre-trained on a large amount of EHR data and then fine-tuned for specific downstream tasks~\cite{rasmy2021med, pang2021cehr, li2020behrt, shang2019pre}. 

We use BERT as the model $f_{\theta}$ to estimate propensity scores using claims records based on our problem setup. Specifically, we use two different schemes to construct input representations for BERT. In the first scheme, we construct an input representation by organizing code representations in chronological order. The code representations are generated using an embedding layer, and positional encodings are added to the code representations to incorporate positional information. The codes from the $t$-th record are added with the corresponding positional encodings based on their positions. Figure~\ref{fig:BERT_code} provides an illustration of how BERT is utilized for estimating propensity scores using claims records with the first input representation scheme. This input representation scheme is commonly employed in existing studies that leverage BERT for modeling EHRs~\cite{rasmy2021med,li2020behrt}. In the second scheme, we apply average pooling to obtain record representations, which is similar to the approach used in LSTM. Subsequently, the record representations are organized chronologically and added with positional encodings based on their positions within the order. Figure~\ref{fig:BERT_record} visually depicts the use of BERT for estimating propensity scores using claims records with the second input representation scheme.

We refer to BERT with the first input representation scheme as $\text{BERT}_{\rm code}$ and BERT with the second input representation scheme as $\text{BERT}_{\rm record}$. We use positional encodings based on sine and cosine functions for both schemes, following the original publications~\cite{vaswani2017attention,devlin2019bert}. In both $\text{BERT}_{\rm code}$ and $\text{BERT}_{\rm record}$, we place \texttt{[CLS]} token at the first position of every input representation and use the learned \texttt{[CLS]} token representation to estimate propensity score. A fully-connected neural network layer with sigmoid activation is used to transform the final \texttt{[CLS]} token representation to the estimated propensity score. The model is trained by minimizing the cross-entropy loss between the estimated propensity score and the binary treatment assignment.

\section{Experiments}

\subsection{Synthetic Dataset}

We formulate a synthetic dataset to conduct controlled experiments with potential time-dependent confounding scenarios based on our problem setup. For notational simplicity, we omit the index for each sample. First, we specify the number of samples $N$ in the dataset, the number of records $T$ for each sample, and the dimensionality of the records $d_x$. We set $N=12000$ and $d_x = 100$. Each sample is independently generated and contains $T$ total records, where $T\sim\text{Poisson}(\lambda=10)$. 

Second, we generate static and dynamic variables for each sample. The static variable matrix $\mathbf{B} \in \mathbb{R}^{T \times d_x} $ represents intrinsic characteristics of the sample that influence health status and remain constant over time. Examples of static variables include gender and genetic characteristics. The dynamic variable matrix $\mathbf{C} \in \mathbb{R}^{T \times d_x}$ represents factors that influence health status and can change over time. Examples of dynamic variables include immune system states and socioeconomic status. Both types of variables are independently generated for each sample and are used as parameters to generate the claims records of the sample.

Third, we generate claims records for each sample using the both static and dynamic variables. The occurrence probability of the $k$-th dimension at time point $t$ is generated from a Beta distribution using the corresponding static and dynamic variables as follows:
\begin{equation*}
    \mathbf{P}_{tk} \sim \text{Beta}(\mathbf{B}_{tk},\mathbf{C}_{tk}),  
\end{equation*}
where $\mathbf{P}_{tk}$, $\mathbf{B}_{tk}$, and $\mathbf{C}_{tk}$ are the $(t,k)$ entry of occurrence probability matrix $\mathbf{P} \in [0,1]^{T \times d_x}$, the static variable matrix $\mathbf{B}$, and the dynamic variable matrix $\mathbf{C}$, respectively. Subsequently, the claims record $\mathbf{X} \in \{0,1\}^{T \times d_x}$ is generated by using the Bernoulli distribution as follows:
\begin{equation*}
    \mathbf{X}_{tk} \sim \text{Bernoulli}(\mathbf{P}_{tk}),
\end{equation*}
where $\mathbf{X}_{tk}$ and $\mathbf{P}_{tk}$ represent $(t,k)$ entry of the claims record $\mathbf{X}$ and occurrence probability matrix $\mathbf{P}$, respectively. 

Based on the synthetic dataset generated using the aforementioned process, we design three confounding scenarios: \textit{consecutive occurrence}, \textit{occurrence distance}, and \textit{occurrence window}. In the \textit{consecutive occurrence} scenario, higher true propensity scores and outcomes are given to samples with consecutive occurrences of a specific code. The \textit{occurrence distance} scenario allocates higher true propensity scores and outcomes to samples with shorter record-wise distances between the occurrences of two specific codes. In the \textit{occurrence window} scenario, samples with more occurrences within a specific lookup window receive higher true propensity scores and outcomes. 

Treatment is assigned by using the Bernoulli distribution with the true propensity score of a sample as $A \sim \text{Bernoulli} \left ( e(X) \right )$. We assume homogeneous and additive treatment effect for all samples. In order to accurately estimate propensity score under all three scenarios, the model needs to learn the temporal patterns of the specific codes that influence the true propensity score. Additional details about the generation of the synthetic dataset and the confounding scenarios are available in Appendix~\ref{appendix:dataset}.

\subsection{Semi-synthetic Dataset}

We formulate a semi-synthetic dataset using Synthea~\cite{walonoski2018synthea} to simulate a potential real-world time-dependent confounding scenario. Synthea is a simulated dataset derived from real-world EHRs~\cite{walonoski2018synthea}. We consider a hypothetical scenario where we aim to estimate the treatment effect for \textit{viral sinusitis} and modify Synthea to introduce time-dependent confounding as follows: we identify two disease codes, \textit{chronic sinusitis} and \textit{viral sinusitis}; and assign higher true propensity scores and outcomes to samples with shorter record-wise distances between occurrences of these two codes. We chose \textit{chronic sinusitis} and \textit{viral sinusitis} to create confounding due to their similar symptoms but differing treatments, potentially leading to confounding when estimating the treatment effect of a medication for \textit{viral sinusitis} in the presence of \textit{chronic sinusitis}~\cite{battisti2023sinusitis}. Treatment is assigned using the Bernoulli distribution with the true propensity score of a sample similar to the synthetic data. We also assume homogeneous and additive treatment effect for all samples. Additional details about the semi-synthetic dataset are available in Appendix~\ref{appendix:dataset}.

\subsection{Experiment Setup}

\paragraph{Baseline methods} We use logistic regression and multi-layer perceptrons (MLP) as baseline methods, which have been commonly used for estimating propensity scores. To construct the covariate vector for each sample, we sum the occurrences of each code across all records, resulting in a vector where the $k$-th element represents the occurrence count of the $k$-th code. Subsequently, we standardize the covariate vectors using the mean and standard deviation of occurrences per code calculated from the entire dataset. In addition, we employ high-dimensional propensity score adjustment (HDPS)~\cite{schneeweiss2009high} as the feature processing method for the baselines. HDPS is widely used in propensity score methods and adds proxy variables to adjust the confounding in longitudinal EHRs with a large number of variables~\cite{schneeweiss2009high}.

\paragraph{Evaluation metrics}

We use mean absolute error (MAE) of the propensity score and ATE as evaluation metrics. The MAE of the propensity score is calculated by averaging the absolute errors between the estimated and the true propensity scores on the samples in the evaluation set, which is defined as follows:
\begin{equation}
    \frac{1}{M} \sum_{i=1}^{M} |e^{(i)} - \hat{e}^{(i)}|,
\end{equation}
where $e^{(i)}$ and $\hat{e}^{(i)}$ are the true and estimated propensity score of sample $i$, respectively, and $M$ is the sample size of the evaluation set. We also use a weighted MAE of the propensity score, which is defined as follows:
\begin{equation}
     \frac{1}{M} \sum_{i=1}^{M} w^{(i)} |e^{(i)} - \hat{e}^{(i)}|,
\end{equation}
where $w^{(i)}$ is the weight corresponding to the error of the estimation for sample $i$. We use the true propensity score as the weight by setting $w^{(i)}=e^{(i)}$, placing more importance to samples with severe confounding in the evaluation.

The MAE of ATE is the absolute error between the estimated and true treatment effect, calculated as $|\hat{\Delta}_{\text{IPTW}} - \Delta_{\text{True}}|$. Due to the potential instability of the ATE caused by extreme values of the estimated propensity scores~\cite{stuart2010matching}, we also compute the MAE of ATE after applying symmetric trimming and clipping of the estimated propensity score for evaluation with an adjustment of extreme propensity scores. Symmetric trimming~\cite{crump2009dealing} excludes samples from the calculation of the MAE of ATE if their estimated propensity scores fall outside the range $[\alpha, 1-\alpha]$, where $\alpha$ is a pre-defined threshold. Symmetric clipping clips the estimated propensity scores with a minimum value of $\alpha$ and a maximum value of $(1-\alpha)$ when computing the MAE of ATE.

\paragraph{Implementation details} We evaluate the models and baselines using 10-fold cross-validation on the synthetic dataset and 5-fold cross-validation on the semi-synthetic dataset. Optimal hyperparameters are selected by using a validation set and a subset of training set. Optimal hyperparameters are selected by using a validation set and a subset of training set. We use Adam~\cite{KingBa15} optimizer without learning rate scheduling. Additional details on hyperparameter selection and implementation for the models and baselines can be found in Appendix~\ref{appendix:hyperparameters}. The code for the experiments is available at ~\url{github.com/Jayaos/propensity_score_dl}.

\subsection{Results}

\begin{table}[htbp]
\footnotesize
\centering
\caption{Experiment results using the synthetic dataset with three confounding scenarios. PS denotes the MAE of the propensity score. $\text{PS}_{\text{\rm W}}$ represents the weighted MAE of the propensity score. We present the average values with their 95\% confidence intervals, which are calculated using 10-fold cross-validation. 
$\text{ATE}_{\text{\rm T}}$ and $\text{ATE}_{\text{\rm C}}$ denote ATE with symmetric trimming and clipping, respectively. LR and MLP indicate logistic regression and multi-layer perceptrons, respectively. LR-HDPS and MLP-HDPS denote  logistic regression and multi-layer perceptrons with high-dimensional propensity score adjustment. Bold values indicate the best performance across all models and baselines.}
\label{tbl:synthetic_dataset_results}
\begin{tabular}{lllllll}
\toprule
Scenario & Model & PS & $\text{PS}_{\text{\rm W}}$ &  ATE &  $\text{ATE}_{\text{\rm T}}$ & $\text{ATE}_{\text{\rm C}}$\\

\midrule
\multirow{7}{*}{\textit{Consecutive Occurrence}}
& LSTM & $\textbf{0.147}_{\pm .005}$ & $0.076_{\pm .003}$  & $5.52_{\pm 1.24}$ & $5.56_{\pm 1.30}$ & $5.74_{\pm 1.24}$ \\
& $\text{BERT}_{\rm code}$  & $0.159_{\pm .003}$ & $\textbf{0.074}_{\pm .001}$ & $\textbf{3.64}_{\pm 1.75}$ & $\textbf{3.64}_{\pm 1.75}$ & $\textbf{3.64}_{\pm 1.75}$ \\
& $\text{BERT}_{\rm record}$  & $0.162_{\pm .003}$ & $0.083_{\pm .004}$ & $6.71_{\pm 1.18}$ & $6.70_{\pm 1.18}$ & $6.71_{\pm 1.18}$ \\
& MLP	& $0.160_{\pm .004}$ & $0.085_{\pm .003}$& $7.18_{\pm .830}$ & $7.23_{\pm .829}$ & $7.16_{\pm .890}$ \\ 
& MLP-HDPS	& $0.186_{\pm .005}$ & $0.097_{\pm .003}$ & $8.53_{\pm 1.71}$ & $8.52_{\pm 1.75}$ & $8.51_{\pm 1.76}$ \\
& LR &  $0.156_{\pm .002}$ & $0.085_{\pm .002}$ & $7.53_{\pm 1.13}$ & $7.53_{\pm1.13}$ & $7.53_{\pm 1.13}$  \\
& LR-HDPS &  $0.184_{\pm .003}$  & $0.100_{\pm .003}$ & $9.44_{\pm .795}$ & $9.44_{\pm .795}$ & $9.44_{\pm .795}$ \\ 

\hline
\multirow{7}{*}{\textit{Occurrence Distance}}
& LSTM & $\textbf{0.079}_{\pm .003}$ & $\textbf{0.033}_{\pm .003}$ & $\textbf{2.19}_{\pm .983}$ & $\textbf{2.14}_{\pm .983}$ & $\textbf{2.10}_{\pm 1.00}$ \\
& $\text{BERT}_{\rm code}$  & $0.100_{\pm .007}$ & $0.051_{\pm .002}$ & $2.69_{\pm 1.99}$ & $2.70_{\pm 2.00}$ & $2.69_{\pm 2.00}$ \\
& $\text{BERT}_{\rm record}$  & $0.113_{\pm .004}$ & $0.055_{\pm .002}$ & $3.69_{\pm 1.64}$ & $3.68_{\pm 1.63}$ & $3.69_{\pm 1.65}$ \\
& MLP	& $0.118_{\pm .002}$ & $0.057_{\pm .001}$ & $4.20_{\pm 1.20}$ & $4.21_{\pm 1.20}$ & $4.20_{\pm 1.19}$ \\ 
& MLP-HDPS	& $0.123_{\pm .002}$ & $0.058_{\pm .002}$ & $4.65_{\pm 1.29}$ & $4.64_{\pm 1.33}$ & $4.63_{\pm 1.29}$ \\
& LR & $0.118_{\pm .002}$ & $0.057_{\pm .001}$ & $4.45_{\pm .991}$ & $4.45_{\pm .991}$ & $4.45_{\pm .991}$ \\ 
& LR-HDPS & $0.112_{\pm .007}$ & $0.055_{\pm .002}$ & $3.11_{\pm 1.79}$ & $3.11_{\pm 1.79}$ & $3.11_{\pm 1.79}$ \\

\hline
\multirow{7}{*}{\textit{Occurrence Window}}
& LSTM & $\textbf{0.070}_{\pm .003}$ & $\textbf{0.033}_{\pm .001}$ & $3.36_{\pm 2.16}$ & $\textbf{2.16}_{\pm 1.45}$ & $\textbf{2.27}_{\pm 1.84}$ \\

& $\text{BERT}_{\rm code}$  & $0.153_{\pm .017}$  & $0.048_{\pm .003}$ & $\textbf{2.40}_{\pm 1.63}$ & $2.45_{\pm 1.69}$ & $2.39_{\pm 1.61}$ \\

& $\text{BERT}_{\rm record}$  & $0.112_{\pm .005}$ & $0.057_{\pm .004}$ & $3.39_{\pm 2.57}$ & $3.50_{\pm 2.61}$ & $3.37_{\pm 2.46}$ \\

& MLP	& $0.279_{\pm .001}$ & $0.128_{\pm .004}$ & $10.9_{\pm 1.01}$ & $11.0_{\pm .989}$ & $11.0_{\pm .444}$ \\ 

& MLP-HDPS	& $0.283_{\pm .004}$ & $0.131_{\pm .006}$ & $11.1_{\pm 1.13}$ & $11.1_{\pm 1.17}$ & $11.1_{\pm 1.20}$ \\

& LR & $0.314_{\pm .002}$ & $0.148_{\pm .003}$ & $11.7_{\pm .857}$ & $11.8_{\pm .796}$ & $11.7_{\pm .846}$ \\ 
& LR-HDPS & $0.281_{\pm .003}$ & $0.126_{\pm .008}$ & $10.0_{\pm 1.55}$ & $10.0_{\pm 1.55}$ & $10.0_{\pm 1.55}$  \\

\bottomrule
\end{tabular}
\end{table}

The results on the synthetic dataset with three confounding scenarios are presented in Table~\ref{tbl:synthetic_dataset_results}. Table~\ref{tbl:semisynthetic_dataset_results} shows the results on the semi-synthetic dataset. In both experiments, deep sequence models (LSTM and BERT) outperform the baseline methods in terms of the MAE of the propensity scores and ATE. There were no significant differences observed between the MAE of ATE and MAE of ATE with symmetric trimming and clipping across all models and baselines in both experiments.

The attention weights of the Transformer encoder and decoder can provide explanations of the model~\cite{vaswani2017attention,wiegreffe2019attention}. Therefore, we calculate the average attention weights assigned to confounding variables, all other variables, and \texttt{[CLS]} token to assess whether the attention weights in $\text{BERT}_{\rm record}$ and $\text{BERT}_{\rm code}$ can offer useful information for identifying confounding variables. Since we utilize the final representation corresponding to the \texttt{[CLS]} token to estimate the propensity score, the attention weights from the \texttt{[CLS]} token at the last encoder layer are used to calculate the average attention weights. The results are presented in Table~\ref{tbl:attention_weights}. We observe that the confounding variables receive significantly higher attention weights compared to all other variables and \texttt{[CLS]} token, providing interpretability to the results using $\text{BERT}_{\rm record}$ and $\text{BERT}_{\rm code}$.

\begin{table}[htbp]
\footnotesize
\centering
\caption{Experimental results using the semi-synthetic dataset. We present the average values with 95\% confidence intervals, which are calculated using 5-fold cross-validation.}
\label{tbl:semisynthetic_dataset_results}
\begin{tabular}{llllll}
\toprule
Model & PS & $\text{PS}_{\rm W}$  & ATE &  $\text{ATE}_{\rm T}$ & $\text{ATE}_{\rm C}$\\

\midrule
LSTM & $\textbf{0.112}_{\pm .021}$ & $\textbf{0.062}_{\pm .014}$ & $0.677_{\pm 1.15}$ & $0.659_{\pm 1.10}$ & $ \textbf{0.496}_{\pm 1.00}$ \\

$\text{BERT}_{\rm code}$  & $0.156_{\pm .016}$  & 
$0.078_{\pm .012}$ & $0.782_{\pm 1.27}$ & $0.769_{\pm 1.25}$ & $0.781_{\pm 1.26}$  \\

$\text{BERT}_{\rm record}$  & $0.156_{\pm .009}$ & $0.077_{\pm .004}$ & $\textbf{0.597}_{\pm 1.05}$ & $\textbf{0.575}_{\pm 1.00}$ & $0.550_{\pm 1.02}$ \\

MLP	& $0.171_{\pm .002}$ & $0.094_{\pm .004}$ & $1.26_{\pm .671}$ & $1.27_{\pm .687}$ & $1.26_{\pm .702}$ \\ 

MLP-HDPS	& $0.171_{\pm .003}$ & $0.094_{\pm .003}$ & $1.11_{\pm .448}$ & $1.07_{\pm .327}$ & $1.07_{\pm .332}$ \\

LR & $0.172_{\pm .003}$ & $0.094_{\pm .004}$ & $1.27_{\pm 1.54}$ & $1.09_{\pm .786}$ & $1.09_{\pm .988}$ \\ 
LR-HDPS & $0.170_{\pm .002}$ & $0.094_{\pm .005}$ & $1.17_{\pm .989}$ & $1.16_{\pm .978}$ & $1.18_{\pm .966}$ \\

\bottomrule
\end{tabular}
\end{table}

\section{Discussion}

Based on the results of two sets of experiments, we demonstrate that deep sequence models (LSTM and BERT) consistently outperform the baselines. This highlights the ability of deep sequence models to capture temporal patterns of confounders, leading to more precise ATE estimation. While LSTM generally performs better than BERT, we conjecture that BERT could show better performance with more complex data and temporal patterns of confounders. This is supported by previous studies that have demonstrated the superiority of BERT over RNNs in various medical tasks using EHRs~\cite{rasmy2021med, pang2021cehr, li2020behrt, shang2019pre}. 

While $\text{BERT}_{\rm code}$ generally shows better performance compared to $\text{BERT}_{\rm record}$, we also believe that $\text{BERT}_{\rm code}$ may encounter challenges when dealing with claims records that have longer histories. Since BERT is known to struggle with learning from long sequences~\cite{devlin2019bert,afkanpour2022bert}, $\text{BERT}_{\rm code}$ might face difficulties in learning from records observed over long periods of time, as the input sequence length of $\text{BERT}_{\rm code}$ increases faster than that of $\text{BERT}_{\rm record}$. We do not observe this issue for the $\text{BERT}_{\rm code}$ in our experiments, possibly because our datasets were not particularly long, with an average number of total codes per sample being less than 40 (Appendix~\ref{appendix:dataset}). However, conducting experiments using claims records containing longer histories, as well as devising efficient representations of a record beyond simple average pooling, remains a subject for future research.

We find that using HDPS for feature processing does not significantly improve the performance of the baseline methods. Our results show that HDPS only slightly improve the performance of logistic regression, as shown in Table~\ref{tbl:synthetic_dataset_results} and Table~\ref{tbl:semisynthetic_dataset_results}. This underscores that the temporal pattern of the confounders cannot be captured readily by rule-based feature processing. Moreover, we observe that extreme values of estimated propensity scores do not occur frequently enough to have a significant impact on the estimation of ATE, as evidenced by the similar values of MAE of ATE with and without symmetric trimming and clipping.

\begin{figure}
    \centering
    
    \begin{subfigure}[b]{0.3\textwidth}
        \centering
        \includegraphics[width=\textwidth]{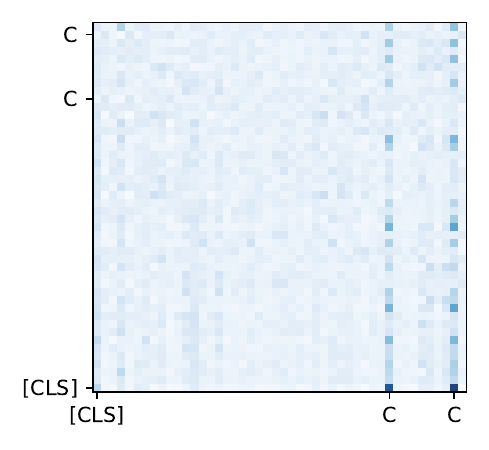}
        \caption{Synthetic dataset-CO}
    \end{subfigure}
    \begin{subfigure}[b]{0.3\textwidth}
        \centering
        \includegraphics[width=\textwidth]{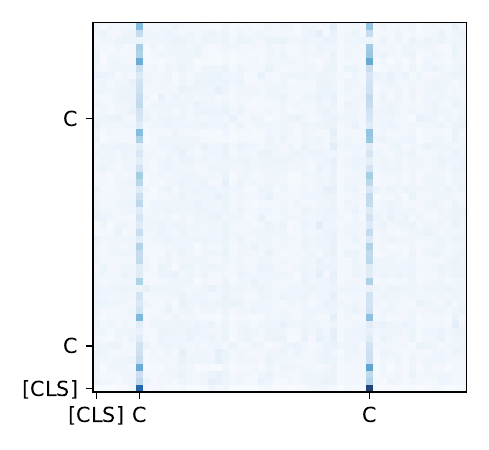}
        \caption{Synthetic dataset-OD}
    \end{subfigure}
    \begin{subfigure}[b]{0.3\textwidth}
        \centering
        \includegraphics[width=\textwidth]{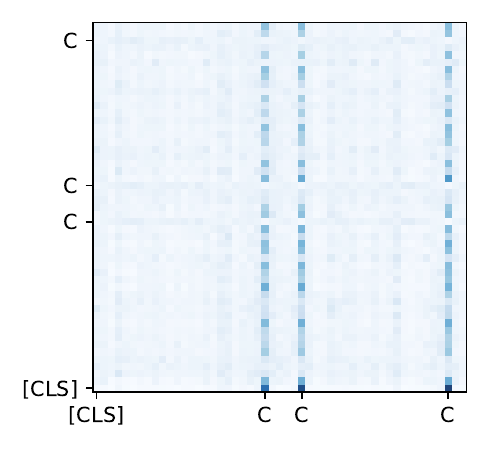}
        \caption{Synthetic dataset-OW}
    \end{subfigure}
    
    \caption{Visualization of attention weights at the last encoder layer of $\text{BERT}_{\rm code}$ for selected samples from the test set. C indicates the position of confounding variables. \texttt{[CLS]} indicates the position of \texttt{[CLS]} token. Darker color represents higher attention weight.}
    \label{fig:attention_weight_visualization}
\end{figure}

We demonstrate the interpretability of BERT by showing that the confounding variables receive significantly higher attention weights than all other variables. Figure~\ref{fig:attention_weight_visualization} provides a visual representation of attention weights at the last encoder layer of $\text{BERT}_{\rm code}$ for selected samples from the test set in each experiment. We observe that \texttt{[CLS]} token pays more attention to the positions of confounding variables. This highlights the utility of BERT in identifying potential confounders in observational data, particularly in cases where prior information about confounders is limited.

\begin{table}[htbp]
\footnotesize
\centering
\caption{The average attention weights assigned to confounding variables, all other variables, and \texttt{[CLS]} token. We present the average values with 95\% confidence intervals calculated using 10-fold cross-validation on the synthetic dataset and 5-fold cross-validation on the semi-synthetic dataset. Synthetic-CO, Synthetic-OD, and Synthetic-OW denote synthetic dataset with \textit{consecutive occurrence}, \textit{occurrence distance}, and \textit{occurrence window} scenario, respectively.}
\label{tbl:attention_weights}
\begin{tabular}{lllllll}
\toprule
\multirow{2}{*}{Dataset} & \multicolumn{2}{c}{Confounding variables} &  \multicolumn{2}{c}{All other variables} & \multicolumn{2}{c}{\texttt{[CLS]} token}\\
 & $\text{BERT}_{\rm code}$  & $\text{BERT}_{\rm record}$ & $\text{BERT}_{\rm code}$ & $\text{BERT}_{\rm record}$ &
 $\text{BERT}_{\rm code}$ &
 $\text{BERT}_{\rm record}$ \\

\hline
Synthetic-CO & 
$0.130_{\pm .014}$ &
 $0.118_{\pm .009}$   &
$0.060_{\pm .005}$  &
 $0.011_{\pm .001}$  &
$0.066_{\pm .015}$ &
$0.042_{\pm .015}$\\
Synthetic-OD & 
$0.276_{\pm .068}$&
$0.200_{\pm .036}$&
$0.013_{\pm .002}$&
$0.062_{\pm .005}$&
$0.022_{\pm .006}$&
$0.055_{\pm .012}$\\
Synthetic-OW &
$0.156_{\pm .037}$&
$0.171_{\pm .040}$&
$0.009_{\pm .002}$&
$0.069_{\pm .007}$&
$0.028_{\pm .007}$&
$0.061_{\pm .018}$\\
\hline
Semi-synthetic &
$0.186_{\pm .034}$&
$0.215_{\pm .060}$&
$0.091_{\pm .016}$&
$0.114_{\pm .030}$&
$0.208_{\pm .025}$&
$0.123_{\pm .037}$\\
\bottomrule
\end{tabular}
\end{table}

\section{Conclusion}

In this study, we empirically demonstrate the effectiveness of deep sequence models in estimating treatment effect on claims records using inverse probability of treatment weighting. Unlike existing methods that require feature processing for propensity score estimation, our study finds that deep sequence models can achieve better performance in estimating propensity score without the need for feature processing. Furthermore, we find that the interpretability of BERT can be used to identify potential confounders even when prior information about confounders is limited, offering practical utility. Based on these findings and results, we believe that inverse probability of treatment weighting using deep sequence models presents a promising approach for treatment effect estimation within the context of our problem setup.

Our study has a few limitations. First, the problem setup for treatment effect estimation is more complex in practice than in our setup. For instance, multiple treatments are often administered rather than a single binary treatment in real-world situations. Moreover, estimating treatment effects over multiple time points would be more practical and useful. Second, our synthetic and semi-synthetic datasets do not fully replicate the complexity of real-world claims records. While we carefully designed multi-step processes with considerations of static and dynamic variables in generating the synthetic dataset, real-world claims records involve countless factors such as disease-disease and drug-disease interactions. Furthermore, our semi-synthetic dataset only considers one hypothetical scenario and may not be a representative example of real-world data.

In future research, it is important to address the limitations identified in this study. This includes developing deep sequence models that can accurately estimate treatment effects in more complex scenarios, such as when multiple treatments are involved or when estimation is needed over multiple time points. Furthermore, applying our methods to real-world claims records will offer valuable insights into the clinical relevance and applicability of our findings.

\begin{ack}
\end{ack}

\newpage

\bibliographystyle{plain}
\bibliography{refs}

\begin{thebibliography}{10}

\bibitem{adler2017electronic}
Julia Adler-Milstein, A~Jay Holmgren, Peter Kralovec, Chantal Worzala, Talisha Searcy, and Vaishali Patel.
\newblock Electronic health record adoption in us hospitals: the emergence of a digital “advanced use” divide.
\newblock {\em Journal of the American Medical Informatics Association}, 24(6):1142--1148, 2017.

\bibitem{afkanpour2022bert}
Arash Afkanpour, Shabir Adeel, Hansenclever Bassani, Arkady Epshteyn, Hongbo Fan, Isaac Jones, Mahan Malihi, Adrian Nauth, Raj Sinha, Sanjana Woonna, et~al.
\newblock Bert for long documents: A case study of automated icd coding.
\newblock In {\em Proceedings of the 13th International Workshop on Health Text Mining and Information Analysis}, pages 100--107, 2022.

\bibitem{austin2015moving}
Peter~C Austin and Elizabeth~A Stuart.
\newblock Moving towards best practice when using inverse probability of treatment weighting (iptw) using the propensity score to estimate causal treatment effects in observational studies.
\newblock {\em Statistics in medicine}, 34(28):3661--3679, 2015.

\bibitem{battisti2023sinusitis}
Amanda~S Battisti, Pranav Modi, and Jon Pangia.
\newblock Sinusitis.
\newblock {\em StatPearls [Internet]}, 2023.

\bibitem{cho2014learning}
Kyunghyun Cho, Bart van Merrienboer, Caglar Guulcehre, Dzmitry Bahdanau, Fethi Bougares, Holger Schwenk, and Yoshua Bengio.
\newblock Learning phrase representations using rnn encoder--decoder for statistical machine translation.
\newblock In {\em Proceedings of the 2014 Conference on Empirical Methods in Natural Language Processing}, pages 1724--1734, 2014.

\bibitem{choi2016doctor}
Edward Choi, Mohammad~Taha Bahadori, Andy Schuetz, Walter~F Stewart, and Jimeng Sun.
\newblock Doctor ai: Predicting clinical events via recurrent neural networks.
\newblock In {\em Machine learning for healthcare conference}, pages 301--318. PMLR, 2016.

\bibitem{crump2009dealing}
Richard~K Crump, V~Joseph Hotz, Guido~W Imbens, and Oscar~A Mitnik.
\newblock Dealing with limited overlap in estimation of average treatment effects.
\newblock {\em Biometrika}, 96(1):187--199, 2009.

\bibitem{devlin2019bert}
Jacob Devlin, Ming-Wei Chang, Kenton Lee, and Kristina Toutanova.
\newblock {BERT}: Pre-training of deep bidirectional transformers for language understanding.
\newblock In {\em Proceedings of the 2019 Conference of the North {A}merican Chapter of the Association for Computational Linguistics}, pages 4171--4186, Minneapolis, Minnesota, June 2019. Association for Computational Linguistics.

\bibitem{evans2016electronic}
R~Scott Evans.
\newblock Electronic health records: then, now, and in the future.
\newblock {\em Yearbook of medical informatics}, 25(S 01):S48--S61, 2016.

\bibitem{graves2013speech}
Alex Graves, Abdel-rahman Mohamed, and Geoffrey Hinton.
\newblock Speech recognition with deep recurrent neural networks.
\newblock In {\em 2013 IEEE international conference on acoustics, speech and signal processing}, pages 6645--6649. Ieee, 2013.

\bibitem{hariton2018randomised}
Eduardo Hariton and Joseph~J Locascio.
\newblock Randomised controlled trials—the gold standard for effectiveness research.
\newblock {\em BJOG: an international journal of obstetrics and gynaecology}, 125(13):1716, 2018.

\bibitem{hirano2003efficient}
Keisuke Hirano, Guido~W Imbens, and Geert Ridder.
\newblock Efficient estimation of average treatment effects using the estimated propensity score.
\newblock {\em Econometrica}, 71(4):1161--1189, 2003.

\bibitem{hochreiter1997long}
Sepp Hochreiter and J{\"u}rgen Schmidhuber.
\newblock Long short-term memory.
\newblock {\em Neural computation}, 9(8):1735--1780, 1997.

\bibitem{jager2008confounding}
KJ~Jager, C~Zoccali, A~Macleod, and FW~Dekker.
\newblock Confounding: what it is and how to deal with it.
\newblock {\em Kidney international}, 73(3):256--260, 2008.

\bibitem{KingBa15}
Diederik Kingma and Jimmy Ba.
\newblock Adam: A method for stochastic optimization.
\newblock In {\em International Conference on Learning Representations)}, 2015.

\bibitem{lee2021severity}
Junghwan Lee, Casey Ta, Jae~Hyun Kim, Cong Liu, and Chunhua Weng.
\newblock Severity prediction for covid-19 patients via recurrent neural networks.
\newblock {\em AMIA Summits on Translational Science Proceedings}, 2021:374, 2021.

\bibitem{li2020behrt}
Yikuan Li, Shishir Rao, Jos{\'e} Roberto~Ayala Solares, Abdelaali Hassaine, Rema Ramakrishnan, Dexter Canoy, Yajie Zhu, Kazem Rahimi, and Gholamreza Salimi-Khorshidi.
\newblock Behrt: transformer for electronic health records.
\newblock {\em Scientific reports}, 10(1):7155, 2020.

\bibitem{lipton2015learning}
Zachary~Chase Lipton, David~C. Kale, Charles Elkan, and Randall~C. Wetzel.
\newblock Learning to diagnose with {LSTM} recurrent neural networks.
\newblock In {\em International Conference on Learning Representations}, 2016.

\bibitem{lunceford2004stratification}
Jared~K Lunceford and Marie Davidian.
\newblock Stratification and weighting via the propensity score in estimation of causal treatment effects: a comparative study.
\newblock {\em Statistics in medicine}, 23(19):2937--2960, 2004.

\bibitem{mcdermott2021comprehensive}
Matthew McDermott, Bret Nestor, Evan Kim, Wancong Zhang, Anna Goldenberg, Peter Szolovits, and Marzyeh Ghassemi.
\newblock A comprehensive ehr timeseries pre-training benchmark.
\newblock In {\em Proceedings of the Conference on Health, Inference, and Learning}, pages 257--278, 2021.

\bibitem{olier2023causal}
Ivan Olier, Yiqiang Zhan, Xiaoyu Liang, and Victor Volovici.
\newblock Causal inference and observational data.
\newblock {\em BMC Medical Research Methodology}, 23(1):227, 2023.

\bibitem{pang2021cehr}
Chao Pang, Xinzhuo Jiang, Krishna~S Kalluri, Matthew Spotnitz, RuiJun Chen, Adler Perotte, and Karthik Natarajan.
\newblock Cehr-bert: Incorporating temporal information from structured ehr data to improve prediction tasks.
\newblock In {\em Machine Learning for Health}, pages 239--260. PMLR, 2021.

\bibitem{radford2019language}
Alec Radford, Jeffrey Wu, Rewon Child, David Luan, Dario Amodei, Ilya Sutskever, et~al.
\newblock Language models are unsupervised multitask learners.
\newblock {\em OpenAI blog}, 1(8):9, 2019.

\bibitem{rasmy2021med}
Laila Rasmy, Yang Xiang, Ziqian Xie, Cui Tao, and Degui Zhi.
\newblock Med-bert: pretrained contextualized embeddings on large-scale structured electronic health records for disease prediction.
\newblock {\em NPJ digital medicine}, 4(1):86, 2021.

\bibitem{rosenbaum1983central}
Paul~R Rosenbaum and Donald~B Rubin.
\newblock The central role of the propensity score in observational studies for causal effects.
\newblock {\em Biometrika}, 70(1):41--55, 1983.

\bibitem{ross2021veridical}
Ryan~D Ross, Xu~Shi, Megan~EV Caram, Phoebe~A Tsao, Paul Lin, Amy Bohnert, Min Zhang, and Bhramar Mukherjee.
\newblock Veridical causal inference using propensity score methods for comparative effectiveness research with medical claims.
\newblock {\em Health Services and Outcomes Research Methodology}, 21(2):206--228, 2021.

\bibitem{rubin1974estimating}
Donald~B Rubin.
\newblock Estimating causal effects of treatments in randomized and nonrandomized studies.
\newblock {\em Journal of educational Psychology}, 66(5):688, 1974.

\bibitem{rubin2005causal}
Donald~B Rubin.
\newblock Causal inference using potential outcomes: Design, modeling, decisions.
\newblock {\em Journal of the American Statistical Association}, 100(469):322--331, 2005.

\bibitem{schneeweiss2009high}
Sebastian Schneeweiss, Jeremy~A Rassen, Robert~J Glynn, Jerry Avorn, Helen Mogun, and M~Alan Brookhart.
\newblock High-dimensional propensity score adjustment in studies of treatment effects using health care claims data.
\newblock {\em Epidemiology}, 20(4):512--522, 2009.

\bibitem{schulam2017reliable}
Peter Schulam and Suchi Saria.
\newblock Reliable decision support using counterfactual models.
\newblock {\em Advances in neural information processing systems}, 30, 2017.

\bibitem{seeger2005application}
John~D Seeger, Paige~L Williams, and Alexander~M Walker.
\newblock An application of propensity score matching using claims data.
\newblock {\em Pharmacoepidemiology and drug safety}, 14(7):465--476, 2005.

\bibitem{shang2019pre}
Junyuan Shang, Tengfei Ma, Cao Xiao, and Jimeng Sun.
\newblock Pre-training of graph augmented transformers for medication recommendation.
\newblock In {\em 28th International Joint Conference on Artificial Intelligence}, pages 5953--5959. International Joint Conferences on Artificial Intelligence, 2019.

\bibitem{shi2022learning}
Jingpu Shi and Beau Norgeot.
\newblock Learning causal effects from observational data in healthcare: a review and summary.
\newblock {\em Frontiers in Medicine}, 9:864882, 2022.

\bibitem{sibbald1998understanding}
Bonnie Sibbald and Martin Roland.
\newblock Understanding controlled trials. why are randomised controlled trials important?
\newblock {\em BMJ: British Medical Journal}, 316(7126):201, 1998.

\bibitem{siebra2024transformers}
Clauirton~A Siebra, Mascha Kurpicz-Briki, and Katarzyna Wac.
\newblock Transformers in health: a systematic review on architectures for longitudinal data analysis.
\newblock {\em Artificial Intelligence Review}, 57(2):1--39, 2024.

\bibitem{soni2019comparison}
Payal~D Soni, Holly~E Hartman, Robert~T Dess, Ahmed Abugharib, Steven~G Allen, Felix~Y Feng, Anthony~L Zietman, Reshma Jagsi, Matthew~J Schipper, and Daniel~E Spratt.
\newblock Comparison of population-based observational studies with randomized trials in oncology.
\newblock {\em Journal of clinical oncology}, 37(14):1209, 2019.

\bibitem{stuart2010matching}
Elizabeth~A Stuart.
\newblock Matching methods for causal inference: A review and a look forward.
\newblock {\em Statistical science: a review journal of the Institute of Mathematical Statistics}, 25(1):1, 2010.

\bibitem{sutskever2014sequence}
Ilya Sutskever, Oriol Vinyals, and Quoc~V Le.
\newblock Sequence to sequence learning with neural networks.
\newblock {\em Advances in neural information processing systems}, 27, 2014.

\bibitem{vaswani2017attention}
Ashish Vaswani, Noam Shazeer, Niki Parmar, Jakob Uszkoreit, Llion Jones, Aidan~N Gomez, {\L}ukasz Kaiser, and Illia Polosukhin.
\newblock Attention is all you need.
\newblock {\em Advances in neural information processing systems}, 30, 2017.

\bibitem{walonoski2018synthea}
Jason Walonoski, Mark Kramer, Joseph Nichols, Andre Quina, Chris Moesel, Dylan Hall, Carlton Duffett, Kudakwashe Dube, Thomas Gallagher, and Scott McLachlan.
\newblock Synthea: An approach, method, and software mechanism for generating synthetic patients and the synthetic electronic health care record.
\newblock {\em Journal of the American Medical Informatics Association}, 25(3):230--238, 2018.

\bibitem{wiegreffe2019attention}
Sarah Wiegreffe and Yuval Pinter.
\newblock Attention is not not explanation.
\newblock In {\em 2019 Conference on Empirical Methods in Natural Language Processing}, pages 11--20. Association for Computational Linguistics, 2019.

\bibitem{xiao2018opportunities}
Cao Xiao, Edward Choi, and Jimeng Sun.
\newblock Opportunities and challenges in developing deep learning models using electronic health records data: a systematic review.
\newblock {\em Journal of the American Medical Informatics Association}, 25(10):1419--1428, 2018.

\bibitem{zeng2022uncovering}
Jiaming Zeng, Michael~F Gensheimer, Daniel~L Rubin, Susan Athey, and Ross~D Shachter.
\newblock Uncovering interpretable potential confounders in electronic medical records.
\newblock {\em Nature communications}, 13(1):1014, 2022.

\end{thebibliography}

\newpage

\appendix
\counterwithin{figure}{section}
\counterwithin{table}{section}

\section{Additional Dataset Details}
\label{appendix:dataset}

\subsection{Synthetic Dataset}

The static variable for each dimension is generated from a uniform distribution on the interval $[5,10]$, resulting in a static variable vector $\mathbf{b} \in [5,10]^{d_x}$. Since static variables do not change over time, $\mathbf{B}$ is constructed as a matrix with $T$ identical rows of $b$.  For the dynamic variables, we generate dynamic variable vector $\mathbf{c} \in [240,260]^{d_x}$, then construct a matrix $\mathbf{C}$ with $T$ identical rows of $\mathbf{c}$. To incorporate the time-varying property of dynamic variables, we use $\text{B-spline}(t)$ sampled from the mixture of five quartic splines to generate $T$-dimensional time-varying coefficient vector for $k$-th dimension as:
\begin{equation*}
    \left ( \text{B-spline}_k(1), \text{B-spline}_k(2), \ldots, \text{B-spline}_k(T) \right ),
\end{equation*}
and then multiply this to $k$-th column of $\mathbf{C}$ (i.e., $\mathbf{C}_{:,k}$). The five quartic splines represent mild incline, mild decline, mild decline after steep incline, mild incline after steep decline, and stable states, respectively. This sampling of B-spline to generate time-varying coefficients is motivated by \cite{schulam2017reliable} and further implementation details can be found in the code at~\url{github.com/Jayaos/propensity_score_dl}.

\paragraph{Consecutive occurrence scenario}

In the \textit{consecutive occurrence} scenario, high propensity scores and outcomes were assigned to samples having consecutive occurrences of a specific code. The true propensity score under the \textit{consecutive occurrence} scenario is assigned as follows:
\begin{equation*}
e(X)=
\begin{cases}
    \sigma (o) + \epsilon & \text{if } o > 1, \\
    0.3 + \epsilon & \text{if } o = 1,\\
    0.1  + \epsilon & \text{if } o = 1,
\end{cases}
\end{equation*}
where $o$ is the maximum number of consecutive occurrences of the specific code and $\sigma (\cdot)$ is a sigmoid function. The noise $\epsilon$ is \textit{i.i.d.} generated from a normal distribution with zero mean and variance of 0.01. The untreated outcome is assigned as follows:
\begin{equation*}
Y_0(X)=
\begin{cases}
    b + \alpha o + \epsilon' & \text{if } o > 1, \\
    b + \epsilon' & \text{if } o \leq 1,\\
\end{cases}
\end{equation*}
where $b$ is a base outcome and $\alpha$ is an outcome coefficient for the maximum number of consecutive occurrences. The noise $\epsilon'$ is \textit{i.i.d.} generated from a normal distribution with zero mean and variance of 0.1. We set $b=10$ and $\alpha=10$. The untreated outcome is assigned as follows:
\begin{equation*}
Y_0(X)=
\begin{cases}
    b + \alpha o + \epsilon' & \text{if } d \geq 1, \\
    b + \epsilon' & \text{if } o \leq 1,\\
\end{cases}
\end{equation*}

\paragraph{Occurrence distance scenario}

The \textit{occurrence distance} scenario assigns high propensity scores and outcomes to samples with shorter record-wise distances between the occurrences of two specific codes. The true propensity score under the \textit{occurrence distance} scenario is assigned as follows:
\begin{equation*}
e(X) = 
\begin{cases}
\sigma ( \log \frac{10}{5d+1}) + \epsilon & \text{if } d \geq 0, \\
0.3 + \epsilon & \text{otherwise} ,
\end{cases}
\end{equation*}
where $d$ is the shortest record-wise distance within the records. The noise $\epsilon$ is \textit{i.i.d.} generated from a normal distribution with zero mean and variance of 0.01. When $d$ cannot be able to be obtained, in the case when one or both of the two specific codes did not occur within the records, we assign $0.3 + \epsilon$. The untreated outcome is assigned as follows:
\begin{equation*}
Y_0(X)=
\begin{cases}
    b + \frac{\alpha}{d+1} + \epsilon' & \text{if } d \geq 0, \\
    b + \epsilon' & \text{otherwise},\\
\end{cases}
\end{equation*}
where $b$ is a base outcome and $\alpha$ is an outcome coefficient for the shortest occurrence distance. The noise $\epsilon'$ is \textit{i.i.d.} generated from a normal distribution with zero mean and variance of 0.1. We set $b=10$ and $\alpha=40$.

\paragraph{Occurrence window scenario}

In the \textit{occurrence window} scenario, samples with more occurrences within a specific lookup window are assigned with high propensity scores and outcomes. We set the lookup window as the last three records of samples. The true propensity score under the \textit{occurrence window} scenario is assigned as follows:
\begin{equation*}
e(X)=
\begin{cases}
    \sigma (c) + \epsilon & \text{if } c > 1, \\
    0.1  + \epsilon & \text{if } c = 0,
\end{cases}
\end{equation*}
where $c$ is the number of occurrences of the code within the lookup window. The untreated outcome is generated as follows:
\begin{equation*}
Y_0(X)=b + \alpha c + \epsilon'
\end{equation*}
where $b$ is a base outcome and $\alpha$ is an outcome coefficient for the occurrence count of the code within the lookup window. The noise $\epsilon'$ is \textit{i.i.d.} generated from a normal distribution with zero mean and variance of 0.1. We set $b=10$ and $\alpha=10$.

\begin{figure}
    \centering
    
    \begin{subfigure}[b]{0.45\textwidth}
        \centering
        \includegraphics[width=\textwidth]{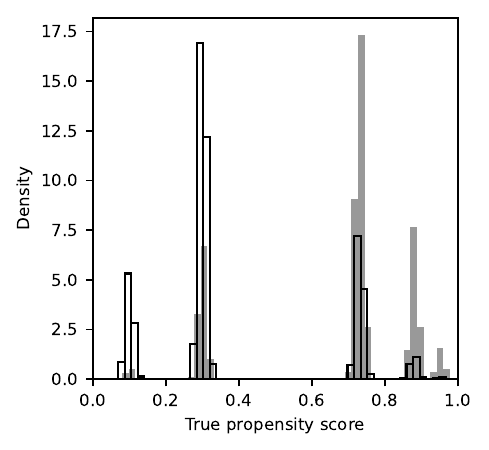}
        \caption{Synthetic dataset, \textit{consecutive occurrence}}
    \end{subfigure}
    \begin{subfigure}[b]{0.45\textwidth}
        \centering
        \includegraphics[width=\textwidth]{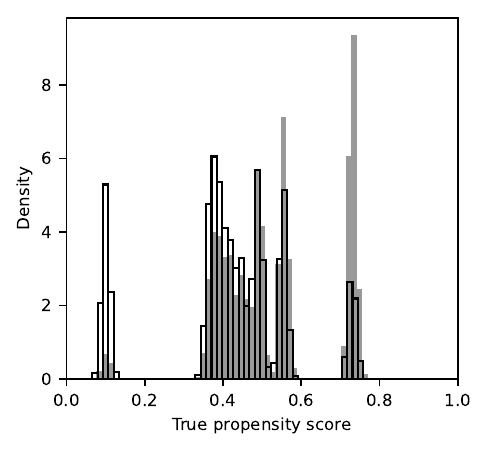}
        \caption{Synthetic dataset, \textit{occurrence distance}}
    \end{subfigure}
    \hfill
    \begin{subfigure}[b]{0.45\textwidth}
        \centering
        \includegraphics[width=\textwidth]{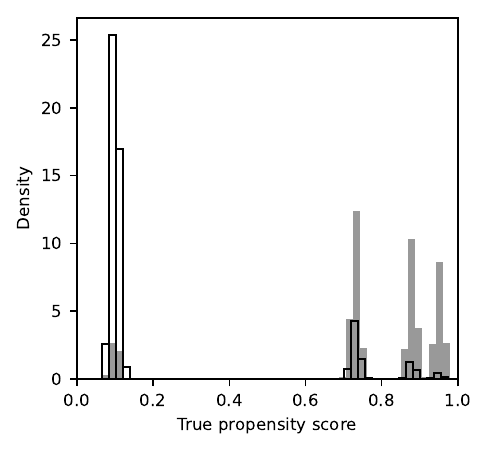}
        \caption{Synthetic dataset, \textit{occurrence window}}
    \end{subfigure}
    \begin{subfigure}[b]{0.45\textwidth}
        \centering
        \includegraphics[width=\textwidth]{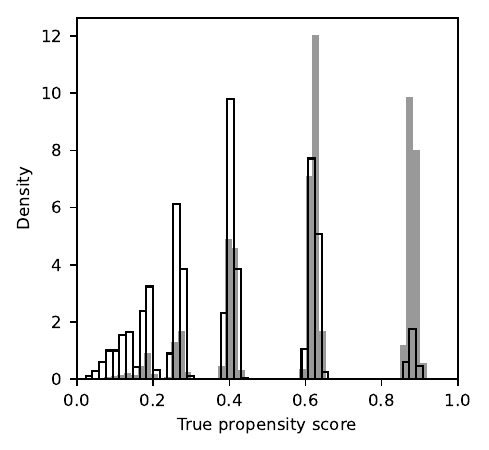}
        \caption{Semi-synthetic dataset}
    \end{subfigure}
    
    \caption{Distributions of propensity scores associated with the synthetic dataset under the three confounding scenarios and the semi-synthetic dataset. The gray bars indicate the treated samples and the unshaded bars indicate the untreated samples.}

\label{fig:ps_distributions}
\end{figure}

Since the samples having meaningfully high confounding rarely occurs in the synthetic dataset, we randomly selected five codes to have increasing probability of occurrences then selected codes for generating confounding among the selected codes. In all scenarios, the treatment effect is identically set to -5, which results in $Y_0 - Y_1 = -5$ for all samples in any confounding scenario. Figure~\ref{fig:ps_distributions} displays the distributions of the true propensity scores associated with the three confounding scenarios. Table~\ref{tbl:dataset_stats} shows the summary statistics of the synthetic dataset with three confounding scenarios.

\subsection{Semi-synthetic Dataset}


With the semi-synthetic dataset, we consider a hypothetical scenario where we aim to estimate the treatment effect for \textit{viral sinusitis} and modified Synthea to introduce time-dependent confounding as follows: we identified two disease codes, \textit{chronic sinusitis} and \textit{viral sinusitis}; and assigned higher true propensity scores and outcomes to samples with a shorter record-wise distance between occurrences of these two codes. This scenario is similar to the \textit{occurrence distance} scenario of the synthetic dataset. The true propensity score is assigned as follows:
\begin{equation*}
e(X)=\sigma (2\log \frac{10}{d^{2.5}}) + \epsilon,
\end{equation*}
where $d$ is the shortest record-wise distance between \textit{chronic sinusitis} and \textit{viral sinusitis}  within the records. The noise $\epsilon$ is \textit{i.i.d.} generated from a normal distribution with zero mean and variance of 0.01. Note that the case when $d=0$ (i.e., \textit{chronic sinusitis} and \textit{viral sinusitis} occur in the same record) was not observed in the dataset. The untreated outcome is generated as follows:
\begin{equation*}
    Y_0(X) = b + \frac{\alpha}{d} + \epsilon',
\end{equation*}
where $b$ is a base outcome and $\alpha$ is an outcome coefficient for the shortest occurrence distance. The noise $\epsilon'$ is \textit{i.i.d.} generated from a normal distribution with zero mean and variance of 0.1. We set $b=10$ and $\alpha=5$. Figure~\ref{fig:ps_distributions} displays the distribution of the true propensity scores. Table~\ref{tbl:dataset_stats} shows the summary statistics of the semi-synthetic dataset.

\begin{table}[htbp]
\footnotesize
\centering
\caption{Summarization statistics of the synthetic and semi-synthetic dataset.}
\begin{tabular}{lccccc}
\toprule
Dataset & size &  avg. record length & avg. codes per sample & avg. codes per record & prev. treated \\

\midrule
Synthetic-CO & 12000 & 11.64 & 40.69 & 3.46 & 0.54 \\
Synthetic-OD & 12000 & 10.10 & 32.00 & 3.18 & 0.47 \\
Synthetic-OW & 12000 & 9.99 & 37.05 & 3.60 & 0.48 \\
\midrule
Semi-synthetic & 6864 & 5.55 & 5.76 & 1.03 & 0.52 \\
\bottomrule
\end{tabular}
\label{tbl:dataset_stats}
\end{table}

\section{Additional Implementation Details}
\label{appendix:hyperparameters}

\subsection{Hyperparameters}

We determined the optimal hyperparameters for all models and baselines by using validation and training sets. Table~\ref{tbl:hyperparameters} shows a list of hyperparameters and their optimal values for all models and baselines. For BERT, the feed-forward dimension was set to 4 times the model dimension. We did not use dropout for LSTM, BERT, and MLP.

\begin{table}[htbp]
\footnotesize
\centering
\caption{List of hyperparameters and their optimal values for all models and baselines.}
\begin{tabular}{ll}
\toprule
 Model & Hyperparameters and their optimal values \\

\midrule
\multirow{5}{*}{LSTM} & Embedding dimension = 64 \\
 & Hidden dimension = 64 \\
 & Number of LSTM layers = 2 \\
 & Learning rate = 0.00001 \\
 & Batch size = 16 \\

\midrule
\multirow{6}{*}{$\text{BERT}_{\rm code}$} & Embedding dimension = 64 \\
 & Model dimension = 64 \\
 & Number of encoder layers = 2 \\
 & Number of attention heads = 4 \\
 & Learning rate = 0.00001 \\
 & Batch size = 16 \\

\midrule
\multirow{6}{*}{$\text{BERT}_{\rm record}$} & Embedding dimension = 64 \\
 & Model dimension = 64 \\
 & Number of encoder layers = 2 \\
 & Number of attention heads = 4 \\
 & Learning rate = 0.00001 \\
 & Batch size = 16 \\

\midrule
\multirow{5}{*}{MLP} & Embedding dimension = 64 \\
 & Number of hidden layers = 2 \\
 & Hidden units = 64 \\
 & Learning rate = 0.0001 \\
 & Batch size = 16 \\

\midrule
\multirow{5}{*}{MLP-HDPS} & Embedding dimension = 64 \\
 & Number of hidden layers = 2 \\
 & Hidden units = 64 \\
 & Learning rate = 0.0001 \\
 & Batch size = 16 \\

\midrule
\multirow{2}{*}{LR} & Learning rate = 0.001 \\
 & Batch size = 16 \\

\midrule
\multirow{2}{*}{LR-HDPS} & Learning rate = 0.001 \\
 & Batch size = 16 \\

\bottomrule
\end{tabular}
\label{tbl:hyperparameters}
\end{table}

\subsection{High-dimensional Propensity Score Adjustment for Baselines Methods}

High-dimensional Propensity Score Adjustment (HDPS) is a multi-step algorithm used to implement high-dimensional proxy adjustment in claims records. In our study, we applied HDPS as a feature processing method for baseline methods (i.e., logistic regression and MLP). While the original publication used a subset of selected covariates based on candidate covariates identification and covariates prioritization~\cite{schneeweiss2009high}, we used all covariates in this study.

HDPS generates 3 new binary covariates per code based on the claims records of a sample: (1) whether the code occurred in the records at least once; (2) whether the code occurred in the records more than the median number of occurrences; and (3) whether the code occurred in the records more than the 75th percentile number of occurrences. For these three covariates, 1 indicates "True" and 0 indicates "False". The median and 75th percentile number of occurrences are calculated from the entire dataset. HDPS results in covariates 3 times the number of unique codes in the dataset.


\end{document}